\documentclass[10pt, a4paper]{article}

\usepackage[final]{lrec-coling2024} 

\usepackage{times}
\usepackage{latexsym}
\usepackage{amsmath}
\usepackage{bm}
\usepackage{graphicx}
\usepackage{booktabs}
\usepackage{float}
\usepackage{multirow}

\makeatletter
\newcommand{\labeltext}[2]{%
  \@bsphack
  \csname phantomsection\endcsname 
  \def\@currentlabel{#1}{\label{#2}}%
  \@esphack
}

\title{CWTM: Leveraging Contextualized Word Embeddings from BERT for Neural Topic Modeling}

\name{Zheng Fang$^1$, Yulan He$^{1,2,3}$ and Rob Procter$^1$$^,$$^3$} 

\address{
$^1$Department of Computer Science, University of Warwick, UK, \\
$^2$Department of Informatics, King's College London, UK, \\
$^3$The Alan Turing Institute, UK \\
         \texttt{\{Z.Fang.4, ~Rob.Procter\}@warwick.ac.uk}\\
         \texttt{yulan.he@kcl.ac.uk} \\}

\abstract{
Most existing topic models rely on bag-of-words (BOW) representation, which limits their ability to capture word order information and leads to challenges with out-of-vocabulary (OOV) words in new documents. Contextualized word embeddings, however, show superiority in word sense disambiguation and effectively address the OOV issue. In this work, we introduce a novel neural topic model called the Contextlized Word Topic Model (CWTM), which integrates contextualized word embeddings from BERT. The model is capable of learning the topic vector of a document without BOW information. In addition, it can also derive the topic vectors for individual words within a document based on their contextualized word embeddings. Experiments across various datasets show that CWTM generates more coherent and meaningful topics compared to  existing topic models, while also accommodating unseen words in newly encountered documents. 
 \\ \newline  }

\begin{document}

\maketitleabstract

\section{Introduction}

Topic modeling has been widely used to explore latent themes within vast document collections, where each document is modeled as a mixture of topics, and a topic is represented by a list of words sorted by their co-occurrence association strength with the topic. Most existing topic models, whether probabilistic based or neural network based, rely on bag-of-words (BOW) information, either as input during training or as the target for training. BOW representations simply encode each document based on word occurrence within the document. It ignores syntactic and semantic relationships between words, consequently limiting the ability of topic models to produce more coherent topics. 

While some topic models incorporate pre-trained word embeddings to capture syntactic and semantic information \cite{nguyen2015improving,li2016topic,zhao2017metalda,dieng2020topic,gupta2019document,gupta2020neural}, they still fall short in capturing the contextual meaning of a word effectively,  as the same words share an identical word embeddings irrespective of their surrounding context. Although some other topic models jointly learn topics and topic-specific word embeddings \cite{shi2017jointly,foulds2018mixed,zhu2020neural} to address this limitation, they do not leverage the advantages of pre-trained language models.

With the development of pre-trained language models like BERT \cite{devlin2018bert}, T5 \cite{raffel2020exploring}, and GPT \cite{radford2019language}, natural language processing (NLP) research has entered a new era. By capturing the surrounding context of each word, these models generate distinct contextualized word embeddings for each occurrence of a word in  text. Contextualized word embeddings are superior to static word embeddings for word sense disambiguation, as they grasp the contextual nuances surrounding word usage. 
They also demonstrate improved capability in managing out-of-vocabulary (OOV) words \cite{akbik2018contextual, han2019unsupervised}. Because contextualized word embeddings are generated contextually, they offer meaningful representations for rare or unseen words. Research has shown that they enhance the performance of various NLP tasks, including text classification, named entity recognition, and question answering \cite{devlin2018bert, raffel2020exploring, radford2019language}.

Topic models can also leverage the advantages of pre-trained language models. The Contextualized Topic Model (CTM) \citep{bianchi-etal-2021-pre} incorporates contextual information from BERT and achieves good performance. However, it only integrates contextual information from document embeddings and still relies on BOW representations as training inputs, thus missing out on the benefits of contextualized word embeddings and suffering from the OOV issue. \citet{thompson2020topic}, \citet{grootendorst2022bertopic} and \citet{meng2022topic} also propose learning topics from pre-trained language models, but they treat topic modeling as a clustering task,  which poses challenges in seamless integration  with other downstream NLP tasks.

In our work, we developed a novel neural topic model combining contextualized word embeddings from BERT \cite{devlin2018bert}. We assume that words with different semantic meanings have different topic information and show that by encoding topic information directly from contextualized word embeddings, our model can produce more coherent and meaningful topics, while also capable of handling unseen words from newly arrived documents. Specifically, we map each contextualized word embedding from BERT to a latent topic vector and aggregate these vectors to represent topics. The document-topic vectors are represented by applying weighted average pooling tp the word-topic vectors in the same document.

In summary, our contributions are three-fold:
\textbf{(1)} We have developed a novel neural topic model called the Contextlized Word Topic Model (CWTM), which integrates contextualized word embeddings from BERT, without relying on the BOW assumption. Our model demonstrates effective mitigation of the OOV issue.
\textbf{(2)} We have compared our model with Latent Dirichlet Allocation (LDA) and other neural topic models across five datasets and shown that it can produce more coherent and meaningful topics.
\textbf{(3)} We have demonstrates that the word-topic vectors learned from our model can improve the performance of downstream tasks such as named entity recognition, indicating that the vectors are semantically meaningful.\footnote{Our code can be found at  \url{https://github.com/Fitz-like-coding/CWTM}.}

\section{Related Work}

Early work has attempted to incorporate pre-trained word embeddings from GloVe \cite{pennington2014glove} and word2vec \cite{mikolov2013distributed} to capture the semantic and syntactic meaning in texts. \citet{nguyen2015improving} integrated word embeddings into conventional topic models LDA \cite{blei2003latent} and GMM \cite{nigam2000text} by replacing the topic-to-word Dirichlet multinomial component in these models by a two-component mixture of a Dirichlet multinomial component and the word embedding component. \citet{li2016topic} combined GMM with word embeddings via the Generalized Po´lya Urn scheme to improve the performance of topic modeling for short texts. \citet{zhao2017metalda} introduced metaLDA incorporating various kinds of document and word meta information including word embeddings information. \citet{gupta2019document} proposed variants of neural autoregressive topic models called DocNADEe and iDocNADEe, using word embeddings as distributional priors, and in 2020 they took a step forward by extending the models to a lifelong neural topic modeling framework \cite{gupta2020neural}.

With the development of large-scale, pre-trained language models, recent work has focused on combining contextualized representations from these models to capture word order information.  \citet{bianchi-etal-2021-pre} introduced the contextualized topic model (CTM) that incorporates document embeddings from BERT to produce more coherent topics. They also introduced Cross-lingual Contextualized Topic Models with Zero-shot Learning to predict topics in different languages \cite{bianchi2020cross}. \citet{mueller2021fine} extended the zero-shot CTM to fine-tune the contextualized document embeddings from the pre-trained language model and observed that it can facilitate cross-linguistic topic modeling. Similarly, \citet{hoyle2020improving} adopted the knowledge distillation technique from \citet{hinton2015distilling} to combine neural topic models with the knowledge learned by the pre-trained language models.

While these studies have made good progress, they only make use of contextual information from document embeddings. None of them incorporates contextualized word embeddings into the topic modeling task. Moreover, they still require bag-of-words (BOW) information as training input or training target. Our work differs in that it incorporates contextualized word embeddings into the topic modeling process, without using any BOW information. Specifically, for each word in a document, our model encodes a word-topic vector from its contextualized word representation.

There has been some work that learns topics from pre-trained language models without using bag-of-words information \cite{thompson2020topic,grootendorst2022bertopic,meng2022topic}. However, they treat topic modeling as an embedding clustering task, and the topic distribution of each document learned does not follow the Dirichlet distribution assumption used in classical topic models, which is not the case for our model.

\begin{figure*}[htb]
\centering
\resizebox{1.0\textwidth}{!}{
\includegraphics[width=1\textwidth]{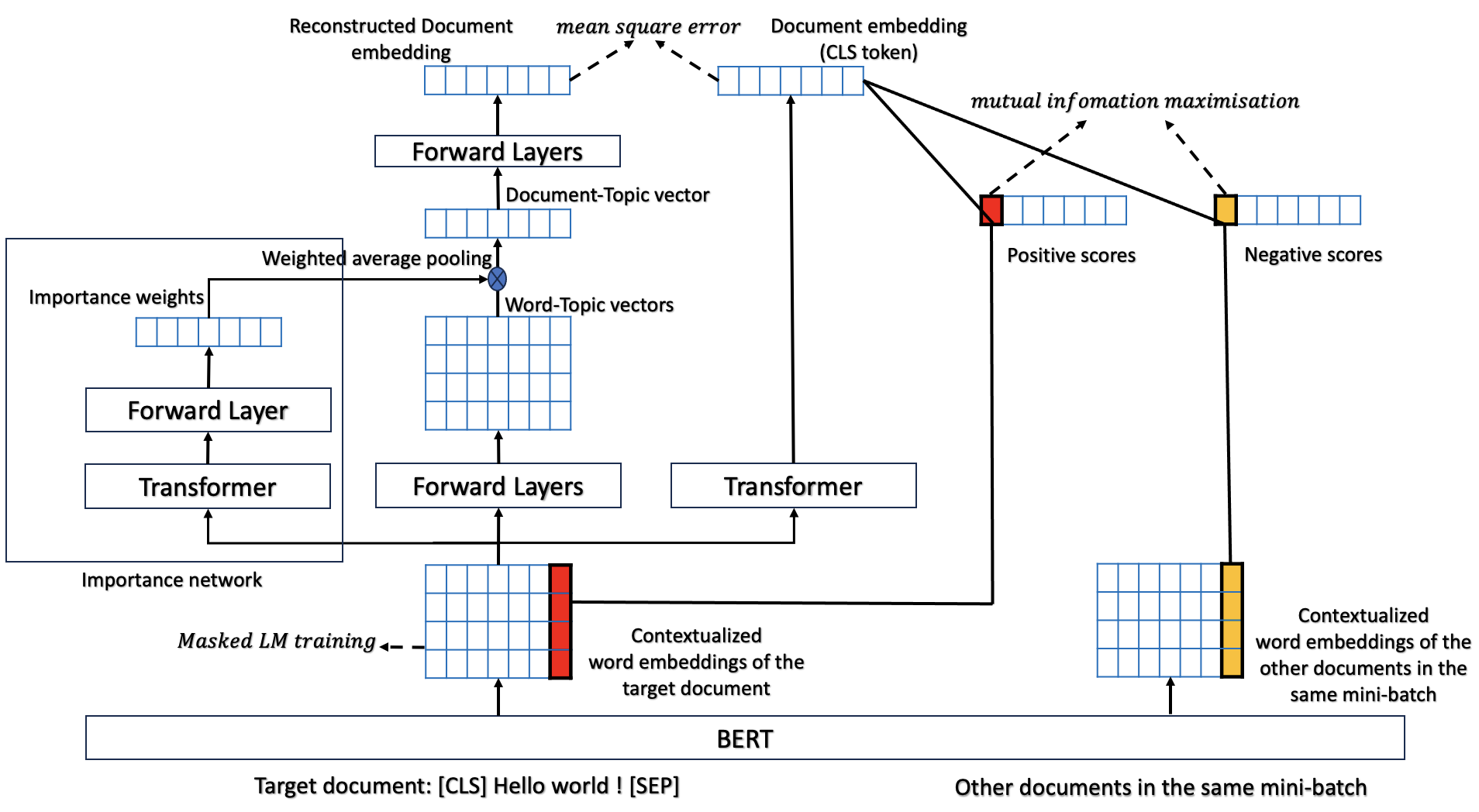}
}
\caption[Model architecture of CWTM.]{
Model architecture. The contextualised word embeddings of the target document are encoded into word-topic vectors, and they are weighted average pooled to generate the document-topic vector, which is regularized to follow the Dirichlet distribution.  The topic vector is then used to reconstruct the document embedding which was learned by a mutual information maximisation objective. A masked language model training objective is also added to regularize the word embeddings.
}
\label{Fig.main1}
\end{figure*}

\section{Contextualized Word Topic Model}

We now introduce our Contextualized Word Topic Model (CWTM). Similar to the WLDA model from \citet{nan2019topic}, CWTM is also based on the Wasserstein autoencoder (WAE) \cite{tolstikhin2017wasserstein}. The WAE is an alternative framework to the variational autoencoder (VAE) \cite{kingma2013auto} for neural topic modeling. It consists of an encoder to map inputs to the latent space and a decoder to reconstruct inputs from the latent space. Similar to VAE, the WAE objective consists of a reconstruction term and a regularization term. However, the regularization term for WAE is designed to minimize the Wasserstein distance between the aggregated posterior and prior distribution, whereas the term for VAE is to minimize the KL divergence between the posterior distribution and prior distribution. \citet{nan2019topic} shown that WAE helps learn better topic representations than VAE-based topic models. 

Different from the WLDA model which learns latent document-topic vectors from BOW representations, CWTM learns latent document-topic vectors from contextualized word embeddings. The encoder of CWTM consists of BERT and a multi-layer perceptron. Each word in a raw text document $d$ is first converted to the contextualized word embedding $e_w$ using BERT. The multi-layer perceptron then maps $e_w$ to a lower-dimensional word-topic vector $\theta_w$. We also construct an importance network consisting of a transformer layer and a single-layer perception followed by the sigmoid activation to map each contextualized word embedding to the corresponding importance weight $\alpha_w$. The document-topic vector $\theta_d$ is then represented as the weighted average pooling of these word-topic vectors.
The decoder consists of a multi-layer perceptron mapping $\theta_d$ back to the document embedding $e_d$ of the document which is learned by a mutual information maximisation strategy. We present the model structure in Figure \ref{Fig.main1}.

\subsection{Combining Contextualized Word Embeddings}
We start by considering learning latent document-topic vectors from document embeddings, which helps us understand the intuition behind combining contextualized word embeddings. We assume document embeddings contain the syntactic and semantic information required for topic information extraction. We could simply use the same WLDA model with document embeddings as input. Let $e_d$ denote the document embedding of a raw text document $d$. The deterministic encoder from WLDA infers the latent document-topic vector: $\theta_d=p(\bm{z}|e_d)=$ encoder($e_d$), where $\bm{z}=[1,2,...,Z]$ are topics and $Z$ is the number of topics pre-specified. The decoder then reconstructs $e_d$ from $\theta_d$. However, in this way, we cannot use decoder matrix weights to represent the topic words of each topic, as we don't have the vocabulary information. To overcome this, we consider learning topic information from contextualized word embeddings. 

Let $e_w$ denote the contextualized word embeddings from $d$, 
we can use a deterministic encoder followed by a softmax function to obtain the latent word-topic vector: $\theta_w=p(\bm{z}|e_w)=$ encoder($e_w$). Knowing $p(\bm{z}|e_w)$ we can then get the topic words of each topic by aggregating the topic-word vectors. We introduce the details in Section \ref{topic_words}. We use weighted average pooling over the latent topic-word vectors to get the document-topic vector $\theta_d$, as different words usually have different importance to the document:
\begin{align}
\theta_d=\sum_{w\in d}\beta_w\theta_w \\\nonumber
\beta_w=\frac{\alpha_w}{\sum_{\hat{w}\in d}\alpha_{\hat{w}}}
\end{align}
where $\alpha_w$ is the importance weight of contextualized word $w$ in the document, which was learned by the importance network we introduced before. We enforce the document-topic vector to follow the Dirichlet distribution and introduce the details in Section \ref{dis_match}.
We then use the document-topic vector to reconstruct the document embedding of the document.


\subsection{Learning Document Embedding}
We utilise mutual information maximisation strategy (MI) \cite{kraskov2004estimating} to learn the document embedding $e_d$. The MI maximizes the mutual information between $e_d$ and the contextualized word embeddings $e_w$ from the same document and minimizes its mutual information with the contextualized word embeddings $e_{\bar{w}}$ from different documents. 


We follow \citet{zhang2020unsupervised} to maximise the mutual information. Instead of using CNN layers to capture the n-gram local contextual dependencies of the input document as in \citet{zhang2020unsupervised}, we add a transformer layer on top of BERT to capture the global contextual dependencies of the input document. We then use the embedding of the [CLS] token from the transformer to represent the document embedding $e_d$. Although there are different ways to represent document embeddings, such as mean pooling, we found no significant difference in the results. 
The document embedding and the contextualized word embeddings from BERT are then used for mutual information computation. The mutual information learning objective of the model is presented below: 
\begin{align}
    L_{MI}(d)=-\log(f(e_w\cdot e_d))-\log(1-f(e_{\bar{w}}\cdot e_d))
\end{align}
where $f$ is the sigmoid activation and $e_{\bar{w}}$ represents each contextualised word embedding from other documents in the same mini-batch. $e_w\cdot e_d$ represents the dot product between them. 

We further add the masked language model (MLM) objective $L_{MLM}$ to regularise the contextualised word embeddings, ensuring that the contextualised information is preserved. The objective becomes:
\begin{align}
    L=L_{MI}(d)+L_{MLM}(d)
\end{align}

\subsection{Distribution Matching}
\label{dis_match}
We enforce the document-topic vector to follow the Dirichlet distribution. We use the distribution matching function \citep{nan2019topic}, enforcing $\theta_d$ to follow Dirichlet distribution via equation \ref{mmd}:
\begin{align}
\label{mmd}
    \widehat{MMD}_{\textbf{k}}(Q_\Theta,P_\Theta)=\frac{1}{m(m-1)}\sum_{i{\neq}j}\textbf{k}({\theta}_i,{\theta}_j)+ \\\nonumber
    \frac{1}{m(m-1)}\sum_{i{\neq}j}\textbf{k}({\theta}_{i}{'},{\theta}_{j}{'}) -\frac{2}{m^2}\sum_{i,j}\textbf{k}(\theta_i,\theta_{j}{'})
\end{align}
where $\widehat{MMD}_{k}(Q_\Theta,P_\Theta)$ is the Maximum Mean Discrepancy (MMD) between the encoded distribution $Q_\theta$ and the prior $P_\theta$. MMD is a distance measure between two distributions \cite{gretton2012kernel}. In our case, $Q_\theta=\theta_d$ and $P_\theta$ is the Dirichlet distribution. $m$ is the size of the mini-batch and $\textbf{k}$ is the information diffusion kernel from \citet{nan2019topic}:
\begin{align}
    \textbf{k}(\theta,\theta^{'})=exp(-arccos^2(\sum_{z=1}^{Z}\sqrt{\theta_z\theta_{z}{'}}))
\end{align}
where $Z$ is the number of topics. 

Instead of only enforcing the document-topic vector to follow the Dirichlet distribution, we also enforce the topic-word vector $\varphi_z$ to follow the Dirichlet distribution. The topic-word vector can be extracted by aggregating the topic vectors of words in the corpus. To reduce the training cost, we extract $\varphi_z$ from each mini-batch and enforce it to follow Dirichlet distribution via $\widehat{MMD}_{k}(Q_\Phi, P_\Phi)$, where $Q_\Phi=\varphi_z$ and $P_\Phi$ is the Dirichlet distribution.

\subsection{Final Objective}
Finally, the decoder consisting of a multi-layer perceptron is used to reconstruct $e_d$ from $\theta_d$, using the mean squared error (MSE) loss:
\begin{align}
    L_{REC}(d)=MSE(e_d,decoder(\theta_d))
\end{align}

We jointly optimise all the objectives so the final objective function is:
\begin{align}
    L=L_{MI}(d)+L_{MLM}(d)+L_{REC}(d)+\\\nonumber
    \widehat{MMD}_{\textbf{k}}(Q_\Theta,P_\Theta)+\widehat{MMD}_{\textbf{k}}(Q_\Phi, P_\Phi)
\end{align}

To maximise the advantages of the pre-trained language model and retain the generalizability learned from pre-training, we also utilized the prompt learning strategy from \citet{jiang2022improved}. Prompts are trainable vectors that allow the optimisation of downstream tasks in an end-to-end manner. During the training, documents are fed into the frozen BERT with the pre-pended soft prompts.

\section{Topic extraction}
\label{topic_words}
Instead of using the decoder matrix weights to extract the topic words of each topic, we extract the topic words of each topic by aggregating the topic vectors of words in the corpus. For example, for all the occurrences of the word “apple” in the corpus, we aggregate their topic vectors with regard to their importance weights in the corresponding documents to represent this word’s weight to each topic. We can then get the topic words of a topic by ranking all words based on the corresponding topic weight. The weight is guaranteed to be a positive value since we applied softmax activation in the encoder. This is similar to the way the Gibbs sampling LDA represents each topic. The difference is that Gibbs LDA assigns a fixed topic to each word, while CWTM assigns a soft topic to each word.


\section{Experiments}
In this section, we introduce the experimental setting of our work, including the datasets and the baseline models used. We focused on two aspects. First, whether our model can produce better quality topics than baseline models. Second, whether our model has a better ability to predict topics in new documents.

\subsection{Datasets}

\begin{table}[H]
\centering
\resizebox{0.48\textwidth}{!}{
\begin{tabular}{|l|l|l|l|l|l|}
\hline
Dataset        & \#train & \#valid & \#vocab & ave.doc.len & \#class \\ \hline
20NG     & 13192  & 5654   & 20794  & 187.4       & 20     \\ \hline
TagMyNews      & 22822  & 9782   & 12328  & 33.1        & 7      \\ \hline
Twitter & 3257   & 1421   & 1730   & 16.1        & 3      \\ \hline
DBpedia        & 560000  & 70000   & 138708  & 46.1        & 14     \\ \hline
AGNews        & 120000  & 7600   & 27521 & 37.8  & 4     \\ \hline
\end{tabular}
}
\caption{Dataset summary}
\label{tab:1}
\end{table}

We tested our model on five datasets, including the 20NG\footnote{https://scikit-learn.org/stable/modules/generated/
sklearn.datasets.fetch\_20newsgroups.html} dataset contains around 18k newsgroup posts on 20 topics; the TagMyNews\footnote{http://acube.di.unipi.it/tmn-dataset/
} dataset contains around 32k short English news from 7 categories; the TweetEval emotion\footnote{https://huggingface.co/datasets/tweet\_eval/viewer/emotion} dataset consists of 4672 tweets from three emotion classes, the DBpedia\footnote{https://huggingface.co/datasets/dbpedia\_14} dataset contains 560k documents from 14 ontology classes, and the AGNews\footnote{https://huggingface.co/datasets/ag\_news} dataset contains 120k news articles from 4 categories.
We summarize the basic statistics of these datasets in Table \ref{tab:1}.

\subsection{Baselines}
We compared our model with four baselines, including LDA \cite{blei2003latent}: A widely used topic model. We implemented the Gibbs Sampling LDA by ourselves\footnote{we will share the code upon paper acceptance}; ProdLDA (\cite{srivastava2017autoencoding}: a neural topic model uses a Gaussian approximation to replace the effect of the Dirichlet prior; WLDA \cite{nan2019topic}: a neural topic model that utilizes Wasserstein distance to regularize the distribution matching; And  CTM \cite{bianchi-etal-2021-pre}: a neural topic model utilizes sentence embeddings from pre-trained language models to enhance the quality of topics. All of these models require bag-of-words representations as the input, whereas CTM requires additional raw text documents as the input. We tuned the hyperparameters of each model and present the settings in Appendix A.

\subsection{Evaluation Metrics}

\begin{table*}[t]
\centering
\resizebox{\textwidth}{!}{%
\begin{tabular}{|c|l|l|l|l|l|l|}
\hline
\multicolumn{1}{|l|}{}     &           & LDA                  & ProdLDA     & WLDA        & CTM         & CWTM                 \\ \hline
\multirow{2}{*}{20NG}      & Coherence & 0.550±0.008          & 0.394±0.010 & 0.371±0.014 & 0.393±0.012 & \textbf{0.600±0.020} \\ \cline{2-7} 
                           & Diversity & 0.630±0.004          & 0.593±0.014 & 0.594±0.003 & 0.604±0.009 & \textbf{0.664±0.005} \\ \hline
\multirow{2}{*}{TagMyNews} & Coherence & 0.586±0.002          & 0.560±0.009 & 0.435±0.021 & 0.508±0.011 & \textbf{0.626±0.016} \\ \cline{2-7} 
                           & Diversity & 0.615±0.003          & 0.620±0.002 & 0.589±0.003 & 0.616±0.002 & \textbf{0.676±0.005} \\ \hline
\multirow{2}{*}{Twitter}   & Coherence & 0.333±0.010          & 0.318±0.001 & 0.310±0.020 & 0.342±0.012 & \textbf{0.357±0.005} \\ \cline{2-7} 
                           & Diversity & 0.495±0.002          & 0.523±0.005 & 0.334±0.023 & 0.514±0.004 & \textbf{0.58±0.0020} \\ \hline
\multirow{2}{*}{DBpedia}   & Coherence & \textbf{0.571±0.012} & 0.538±0.020 & 0.555±0.037 & 0.477±0.014 & 0.569±0.011          \\ \cline{2-7} 
                           & Diversity & 0.649±0.006          & 0.677±0.004 & 0.627±0.001 & 0.655±0.004 & \textbf{0.692±0.004} \\ \hline
\multirow{2}{*}{AGNews}    & Coherence & 0.578±0.003          & 0.550±0.013 & 0.463±0.025 & 0.510±0.003 & \textbf{0.633±0.020} \\ \cline{2-7} 
                           & Diversity & 0.617±0.002          & 0.623±0.005 & 0.575±0.008 & 0.617±0.007 & \textbf{0.680±0.004} \\ \hline
\end{tabular}%
}
\caption{Topic coherence and topic diversity results for 50 topics. We ran each experiment three times and report the average results. We also report the standard deviation.}
\label{tab:50topics}
\end{table*}

We used three metrics to evaluate each model, which are topic coherence, topic diversity and document classification accuracy. We cannot compute ELBO-based perplexity for WLDA and our model because they are not based on variational inference.

\noindent\textbf{Topic coherence} measures the semantic similarity between words within a given topic. It has been widely used in previous work \cite{srivastava2017autoencoding,nan2019topic,bianchi-etal-2021-pre}. We used the best performing measure C\_V from \citet{roder2015exploring}, and used Wikipedia\footnote{https://hobbitdata.informatik.uni-leipzig.de/homes/mroeder/palmetto/Wikipedia\_bd.zip}
as the external corpus. We focused on the top 10 words of each topic as in previous papers \cite{nan2019topic, bianchi-etal-2021-pre} and used the Palmetto library algorithm \cite{roder2015exploring}. 

\noindent\textbf{Topic diversity} assesses the degree of diversity in a set of topics. It has been used as a supplementary measure to topic coherence. We computed the word embeddings-based centroid similarity\footnote{We used the code from OCTIS: https://github.com/MIND-Lab/OCTIS/tree/master} based on the top 10 words of each topic. The diversity score is expressed as 1 minus the similarity score. A higher score indicates more varied topics. For baseline models, we used BERT with the Aggregated Static Embeddings approach from \citet{bommasani2020interpreting} to obtain the static embedding for each topic word. We used the contextualized word embeddings from the first layer of BERT and used mean pooling to distil the static embeddings. \citet{bommasani2020interpreting} found this generally performs better than pre-trained static word embeddings. This also allows us to obtain the embeddings for rare words in the corpus. For CWTM, since it encodes word-topic vectors from contextualized word embeddings, different occurrences of the same word have different weights for each topic. For a fair comparison, a weighted average of the contextualized word embeddings from all occurrences of a word is used to represent its word embeddings for each topic. We also used contextualized word embeddings from the first layer of BERT to maintain consistency with the baseline models. 

\noindent\textbf{Document classification} has also been used in many previous works to asses the predictive performance of the latent document-topic vectors \cite{li2016topic,gui2019neural,nan2019topic}. Higher accuracy indicates better predictive performance. We applied a logistic regression classifier with a default parameters setting. We used the document-topic vectors from the testing set as the input and conducted five-fold cross-validation.

\subsection{Benchmark Results}

\begin{figure}[h!]
\centering
\resizebox{0.5\textwidth}{!}{
\includegraphics[width=1\textwidth]{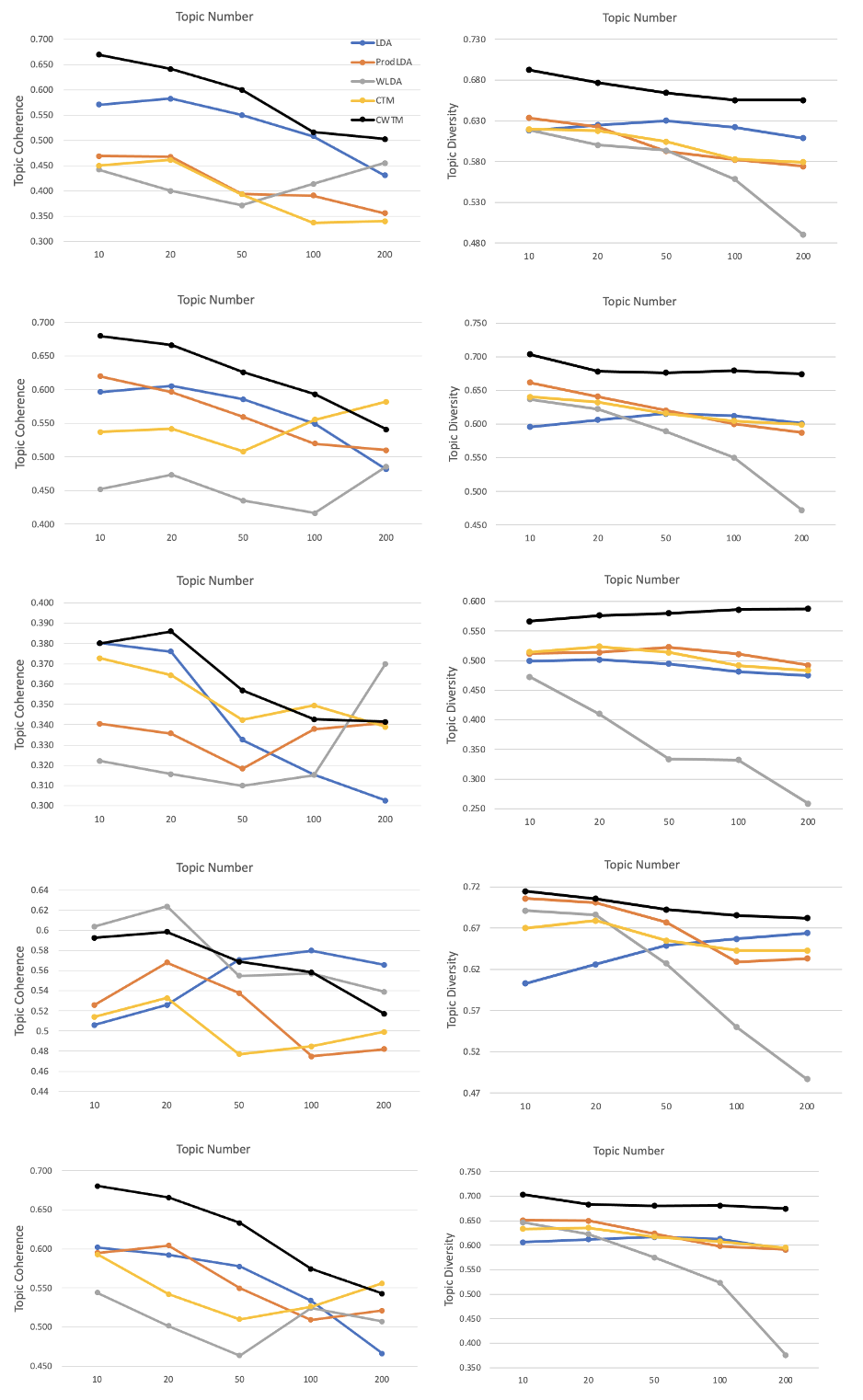}
}
\caption{Topic coherence and topic diversity scores across different numbers of topics by different models for 20NG (top row); TagMyNews (2nd row), Twitter (3rd row), DBpedia (4th row), and AGNews (bottom row).}
\label{Fig.main2}
\end{figure}

\begin{table*}[htb!]
\centering
\resizebox{\textwidth}{!}{%
\begin{tabular}{|c|l|l|l|l|l|l|l|l|}
\hline
\multicolumn{1}{|l|}{}     &                & CWTM                 & w/o $L_{MI}$ & w/o $L_{MLM}$ & w/o $L_{REC}$ & w/o $\widehat{MMD}_{\textbf{k}}(Q_\Theta,P_\Theta)$  & w/o $\widehat{MMD}_{\textbf{k}}(Q_\Phi, P_\Phi)$ & w/o Importance  \\ \hline
\multirow{3}{*}{20NG}      & Coherence      & \textbf{0.641±0.021} & 0.396±0.038   & 0.349±0.028    & 0.514±0.037    & 0.519±0.034          & 0.596±0.005          & 0.517±0.027          \\ \cline{2-9} 
                           & Diversity      & 0.677±0.011          & 0.618±0.009   & 0.000±0.000    & 0.655±0.024    & \textbf{0.690±0.015} & 0.639±0.009          & 0.607±0.002          \\ \cline{2-9} 
                           & Classification & \textbf{0.530±0.024} & 0.206±0.043   & 0.058±0.000    & 0.352±0.048    & 0.511±0.014          & 0.512±0.027          & 0.526±0.012          \\ \hline
\multirow{3}{*}{TagMyNews} & Coherence      & \textbf{0.666±0.018} & 0.566±0.019   & 0.421±0.002    & 0.525±0.043    & 0.597±0.030          & 0.653±0.011          & 0.606±0.021          \\ \cline{2-9} 
                           & Diversity      & 0.678±0.007          & 0.662±0.015   & 0.000±0.000    & 0.663±0.006    & \textbf{0.690±0.006} & 0.639±0.006          & 0.626±0.002          \\ \cline{2-9} 
                           & Classification & 0.819±0.001          & 0.576±0.034   & 0.252±0.000    & 0.626±0.071    & 0.810±0.003          & 0.820±0.001          & \textbf{0.828±0.011} \\ \hline
\multirow{3}{*}{Twitter}   & Coherence      & \textbf{0.386±0.010} & 0.332±0.020   & 0.361±0.035    & 0.378±0.014    & 0.355±0.005          & 0.378±0.017          & 0.368±0.006          \\ \cline{2-9} 
                           & Diversity      & 0.576±0.004          & 0.553±0.029   & 0.021±0.025    & 0.573±0.008    & \textbf{0.592±0.008} & 0.541±0.009          & 0.564±0.007          \\ \cline{2-9} 
                           & Classification & \textbf{0.641±0.013} & 0.472±0.059   & 0.394±0.001    & 0.593±0.027    & 0.610±0.034          & 0.627±0.018          & 0.606±0.020          \\ \hline
\multirow{3}{*}{Dbpeida}   & Coherence      & \textbf{0.599±0.015} & 0.391±0.033   & 0.315±0.009    & 0.519±0.109    & 0.513±0.015          & 0.529±0.017          & 0.509±0.013          \\ \cline{2-9} 
                           & Diversity      & 0.705±0.004          & 0.643±0.016   & 0.000±0.000    & 0.669±0.011    & \textbf{0.718±0.002} & 0.651±0.007          & 0.647±0.025          \\ \cline{2-9} 
                           & Classification & 0.947±0.004          & 0.619±0.029   & 0.079±0.000    & 0.816±0.115    & 0.947±0.013          & \textbf{0.962±0.002} & 0.950±0.009          \\ \hline
\multirow{3}{*}{AGNews}    & Coherence      & \textbf{0.666±0.014} & 0.620±0.010   & 0.390±0.023    & 0.600±0.032    & 0.512±0.026          & 0.661±0.006          & 0.567±0.010          \\ \cline{2-9} 
                           & Diversity      & 0.683±0.003          & 0.675±0.005   & 0.000±0.000    & 0.648±0.016    & \textbf{0.693±0.012} & 0.650±0.006          & 0.632±0.009          \\ \cline{2-9} 
                           & Classification & 0.866±0.005          & 0.745±0.038   & 0.255±0.000    & 0.784±0.031    & 0.867±0.004          & 0.859±0.005          & \textbf{0.874±0.002} \\ \hline
\end{tabular}%
}
\caption{Ablation study of CWTM. CWTM represents our model with all components included.}
\label{tab:ablation}
\end{table*}

We present the topic coherence and topic diversity results of using 50 topics in table \ref{tab:50topics}. It can be observed that CWTM achieved competitive performance compared with baseline models on both topic coherence and topic diversity measures. It achieved the best coherence and diversity scores on the 20NG, TagMyNews, Twitter, and AGNews datasets. It also has the second-best coherence score and best diversity score on the DBpedia dataset. It suggests that encoding topic information from contextualized word embeddings can help generate more coherent and meaningful topics. Although CTM incorporated contextual document embeddings, it did not show significant advantages over other baseline models. One possible reason is that the sentence embedding model used by CTM can only accept up to 128 tokens in the input document, so it cannot handle long documents well. 

We also plot the performance of different models with different numbers of topics settings in Figure \ref{Fig.main2}. We set the number of topics $Z$= \{10, 20, 50, 100, 200\}. It can be observed that overall CWTM consistently outperforms other models in terms of the topic coherence scores on the 20NG, TagMyNews, Twitter, and AGNews datasets. It also has the second-best coherence scores on the DBpedia dataset when the number of topics is below 100.  We can also conclude from the figure that CWTM produces more diverse topics than other baseline models across all numbers of topics settings. We also found that WLDA performed poorly on the topic diversity measure. It's diversity score dropped dramatically when we increased the number of topics.

\begin{table}[h!]
\centering
\resizebox{0.48\textwidth}{!}{%
\begin{tabular}{|l|l|l|l|l|l|}
\hline
        & 20NG                 & TagMyNews            & Twitter              & DBpedia              & AGNews               \\ \hline
LDA     & \textbf{0.589±0.004} & 0.783±0.009          & 0.513±0.009          & 0.864±0.007          & 0.858±0.002          \\ 
ProdLDA & 0.359±0.010           & 0.681±0.003          & 0.437±0.008          & 0.769±0.003          & 0.762±0.008          \\ 
WLDA    & 0.426±0.016          & 0.747±0.007          & 0.415±0.019          & 0.907±0.008          & 0.824±0.014          \\ 
CTM     & 0.494±0.033          & 0.783±0.009          & 0.570±0.021          & 0.881±0.003          & 0.864±0.002          \\ 
CWTM    & 0.530±0.024          & \textbf{0.819±0.001} & \textbf{0.641±0.013} & \textbf{0.947±0.004} & \textbf{0.866±0.005} \\ \hline
\end{tabular}%
}
\caption{Document classification results for different models}
\label{tab:cls}
\end{table}

We also present the document classification results in Table \ref{tab:cls}.  We set the number of topics to 20. It can be observed that CWTM outperforms existing models on four datasets. Only LDA has better classification performance on the 20NG dataset. The excellent classification performance indicates that CWTM can produce document-topic vectors with good predictive power. We also plot the classification performance of different models with different numbers of topics settings in Appendix B. The results are consistent with Table \ref{tab:cls}.





\subsection{Ablation Study}

We study the effect of different objective functions and the importance network in this section. We present the performance of our model without each objective function and the importance network in Table \ref{tab:ablation}. We set the number of topics to 20. It can be observed that with all components included, CWTM achieved the best topic coherence scores over all datasets. By excluding the mutual information loss ($L_{MI}$), masked language model loss ($L_{MLM}$), or reconstruction loss ($L_{REC}$), CWTM's performance on all three evaluation metrics dropped significantly, suggesting that these objectives are critical to the overall performance of the model. By excluding the document level distribution matching objective ($\widehat{MMD}_{\textbf{k}}(Q_\Theta,P_\Theta)$), CWTM produced slightly more diverse but much less coherent topics, indicating that this objective is beneficial for learning coherent topics, while slightly limits the model to learn more diverse topics. Without the topic level distribution matching objective ($\widehat{MMD}_{\textbf{k}}(Q_\Phi, P_\Phi)$), CWTM's topic coherence and diversity scores on all datasets dropped, suggesting that this objective is essential to learn coherent and diverse topics. To exclude the importance network, we treat words in a document as having equal weights, i.e. $\alpha_w = 1 $. We can observe that without the importance network, the coherence and diversity scores dropped significantly, proving that different words have different contributions to a document. We present the topic words of CWTM on different datasets in Appendix C.

\subsection{Handling Out-of-Vocabulary Words}

\begin{table}[H]
\centering
\resizebox{0.48\textwidth}{!}{%
\begin{tabular}{|l|l|l|l|l|l|}
\hline
               & 20NG & TagMyNews & Twitter & DBpedia & AGNews \\ \hline
ave.vocab.Test1     & 29.4       & 16.0      & 4.9    & 17.6  & 18.0  \\ \hline
ave.vocab.Test2 & 56.7        & 15.1       & 7.5      & 20.5 & 18.2   \\ \hline
ave.new\_vocab.Test1 & 1.9        & 0.9      & 0.1    & 1.2 & 1.0   \\ \hline
ave.new\_vocab.Test2 & 27.6        & 5.9      & 3.6    & 8.2 & 7.0   \\ \hline
\#Test1 & 5990        & 6151      & 720    & 1180 & 4589   \\ \hline
\#Test2 & 1252        & 7264      & 1924    & 8192 & 3783   \\ \hline
\end{tabular}%
}
\caption{The vocabulary difference between the Test1 and Test2 datasets. “ave.vocab” represents the average vocabulary size of each document. “ave.new\_vocab” represents the average unobserved vocabulary size in each document. \#test1 and \#test2 represent the number of documents in each test set.}
\label{tab:5}
\end{table}

\begin{table*}[htb!]
\centering
\resizebox{\textwidth}{!}{%
\begin{tabular}{|c|l|l|l|l|l|l|}
\hline
\multicolumn{1}{|l|}{}     &       & LDA         & ProdLDA     & WLDA         & CTM         & CWTM                 \\ \hline
\multirow{3}{*}{20NG}      & $\ge90\%$ & 0.415±0.013 & 0.139±0.008 & 0.166±0.009  & 0.359±0.023 & 0.371±0.017          \\ \cline{2-7} 
                           & $\le70\%$ & 0.293±0.008 & 0.166±0.006 & 0.160±0.008  & 0.310±0.026 & 0.374±0.020          \\ \cline{2-7} 
                           & diff  & 0.122±0.010 & -0.027±0.01 & 0.006±0.013  & 0.050±0.013 & \textbf{-0.004±0.01} \\ \hline
\multirow{3}{*}{TagMyNews} & $\ge90\%$ & 0.708±0.010 & 0.512±0.011 & 0.462±0.029  & 0.718±0.019 & 0.804±0.008          \\ \cline{2-7} 
                           & $\le70\%$ & 0.537±0.009 & 0.307±0.006 & 0.33±0.015   & 0.601±0.025 & 0.721±0.012          \\ \cline{2-7} 
                           & diff  & 0.171±0.003 & 0.206±0.011 & 0.133±0.022  & 0.116±0.01  & \textbf{0.083±0.006} \\ \hline
\multirow{3}{*}{Twitter}   & $\ge90\%$ & 0.489±0.022 & 0.466±0.013 & 0.421±0.016  & 0.55±0.023  & 0.595±0.021          \\ \cline{2-7} 
                           & $\le70\%$ & 0.455±0.010 & 0.443±0.007 & 0.437±0.011  & 0.507±0.021 & 0.589±0.015          \\ \cline{2-7} 
                           & diff  & 0.034±0.031 & 0.023±0.020 & -0.016±0.015 & 0.043±0.018 & \textbf{0.006±0.008} \\ \hline
\multirow{3}{*}{Dbpeida}   & $\ge90\%$ & 0.838±0.011 & 0.678±0.017 & 0.363±0.043  & 0.859±0.007 & 0.827±0.025          \\ \cline{2-7} 
                           & $\le70\%$ & 0.800±0.015 & 0.580±0.012 & 0.294±0.039  & 0.839±0.011 & 0.837±0.020          \\ \cline{2-7} 
                           & diff  & 0.037±0.017 & 0.098±0.025 & 0.069±0.028  & 0.021±0.01  & \textbf{-0.01±0.023} \\ \hline
\multirow{3}{*}{AGNews}    & $\ge90\%$ & 0.826±0.006 & 0.642±0.007 & 0.527±0.027  & 0.831±0.008 & 0.871±0.004          \\ \cline{2-7} 
                           & $\le70\%$ & 0.732±0.005 & 0.479±0.010 & 0.475±0.03   & 0.777±0.008 & 0.85±0.006           \\ \cline{2-7} 
                           & diff  & 0.093±0.007 & 0.163±0.015 & 0.051±0.053  & 0.054±0.015 & \textbf{0.021±0.005} \\ \hline
\end{tabular}%
}
\caption[Document classification accuracy over test sets with different vocabulary size.]{Document classification accuracy over Test1 and Test2. “$\ge90\%$” means each document in Test1 has over 90\% of the vocabulary words observed in the training set. “$\le70\%$” means each document in Test2 has less than 70\% vocabulary words observed in the training set. “diff” shows the accuracy difference between Test1 and Test2. The closer the difference is to 0, the better the performance of handling OOV words. We ran each model 5 times.}
\label{tab:oov_diff}
\end{table*}

We now describe how we compared different models' performance in handling OOV words. For each corpus, we first randomly sampled 1000 documents as the training set. We then sampled two testing sets Test1 and Test2. For each document in Test1, over 90\% of the vocabulary words were observed in the training set, and for each document in Test2, less than 70\% of the vocabulary words were observed in the training set. In other words, documents in Test2 have more OOV words than documents in Test1. We set the number of topics to 20 and trained different models on the training set. We fed the document-topic vectors from the testing sets to a logistic regression classifier and conducted five-fold cross-validation. A model insensitive to OOV words is expected to have consistent classification performance over the two different testing sets. We summarize the basic statistics of the sampled Test1 and Test2 sets in Table \ref{tab:5}.

We report the results in Table \ref{tab:oov_diff}. We observed that CWTM achieved more consistent results over the different testing sets than baseline models. The accuracy difference between CWTM on Test1 and Test2 is smaller than that of the baseline models on all datasets. Student t-tests on accuracy difference show that CWTM outperforms other baselines on almost all datasets, except the WLDA model on the 20NG (t=-1.29, p=0.12) and AGNews (t=-1.28, p=0.12) datasets, ProdLDA model on the Twitter dataset (t=-1.13, p=0.14). However, the higher classification accuracy of CWTM suggests that CWTM is more competitive. This proves that incorporating contextualized word embeddings into topic models can help alleviate the OOV issue.

\subsection{Word-topic Vector Quality}

CWTM can learn latent word-topic vectors for words in a document. To evaluate the predictive performance of the latent word-topic vectors, we used the Named Entity Recognition (NER) metric. We trained CWTM on the conll2003 NER dataset \cite{sang2003introduction} and concatenated the latent vectors with the contextualised word embeddings from BERT's last layer and fed them into a single-layer perceptron for the conll2003 NER task. To learn the latent word-topic vectors using CWTM, we set the latent vector size to 50, the epoch number to 30 and the learning rate to 1e-4. To train the single-layer perceptron, we set the epoch number to 10 and the learning rate to 2e-3. We froze the parameters of BERT and CWTM when feeding the vectors. We used BERT as the baseline and fine-tuned it within 2 epochs using MLM objective on the target corpus. We ran the experiment five times and report the average results. 

\begin{table}[htb]
\centering
\begin{tabular}{|l|l|l|l|}
\hline
                & Precision & Recall & F1    \\ \hline
BERT            & 0.792     & 0.839   & 0.814 \\ \hline
BERT+CWTM & 0.807     & 0.850  & 0.828  \\ \hline
\end{tabular}
\caption{NER performance. The first row shows the results of using only contextualised word embeddings from BERT as the input. The second row shows the results of using concatenated vectors as the input.}
\label{tab:7}
\end{table}

The results are shown in Table \ref{tab:7}. It can be seen that by adding word-topic vectors we can obtain better results for the NER task. A student t-test on f1 scores (t=-3.75, p=0.003) shows this difference was statistically significant. This means that the latent word-topic vectors learned from CWTM are semantically meaningful.

\section{Conclusions}

We have developed a novel neural topic model called CWTM. The model uses the raw text document as input and infers topic information directly from the contextualized word embeddings of the document. In addition to learning the latent document-topic vector of a document, it also learns the latent word-topic vector of each word in the document. Experimental results show that overall CWTM outperforms the baseline models in topic coherence, topic diversity and document classification measures, which means that CWTM can produce more coherent and meaningful topics. The experiments on the OOV words prove that CWTM is more insensitive to unseen words from newly arrived documents.

Experiments with the CoNLL2003 NER dataset further demonstrates the benefits of word-topic vectors learned from CWTM and shows that CWTM can provide new input information for downstream NLP tasks. In the future, we aim to conduct human evaluations to assess the performance of the model on real-world tasks.

\section*{Acknowledgements}
ZF receives the PhD studentship jointly funded by the University of Warwick and China Scholarship Council. YH is supported by a Turing AI Fellowship funded by the UK Research and Innovation (grant no. EP/V020579/2, EP/V020579/2). 

\nocite{*}
\section*{Bibliographical References}\label{sec:reference}

\bibliographystyle{lrec-coling2024-natbib}
\bibliography{lrec-coling2024-example}

\label{lr:ref}
\bibliographystylelanguageresource{lrec-coling2024-natbib}
\bibliographylanguageresource{languageresource}

\clearpage
\appendix
\section*{Appendix A: Parameters Settings \labeltext{A}{sec:appendix_A}}
We trained all models using the training data and report the results for the test data. We used full-length texts as input. To transform documents into BoW representations, each dataset was properly pre-processed with word lemmatizing for each word, and all stop words were removed. We also removed all numeric and non-alphabetic characters, as well as frequently occurring words in each corpus, i.e., the top 10 most frequently occurring words in each corpus. When training CWTM, we did not preprocess the documents because it accepts raw text documents as input. When presenting the topics, we preprocessed the topic words generated by CWTM for a fair comparison with other models.

For Gibbs sampling LDA, we set $\alpha = 1/t$ and $\beta=1/t$, where $t$ is the number of topics for the model and found it outperforms other settings.  We trained it over 2000 iterations. For prodLDA and WLDA, we set the batch number to 256 and trained them over 500 epochs for the 20NG, TagMynews and Twitter datasets. We trained them over 10 epochs for the Dbpedia dataset since the number of documents in Dbpedia is much larger. We found no significant difference in the results compared to training them over 7-9 epochs. We used an ADAM optimizer with high momentum $\beta1 = 0.99$ and a learning rate of 0.002 as it does in \citet{nan2019topic}. We set the \textit{keep\_prob} parameter of ProdLDA to \{0.0, 0.2, 0.4, 0.6\} and only report the best result from them. We set the Dirichlet prior of WLDA to 0.1 which was suggested by the original paper. The Dirichlet noise of WLDA is set to \{0.0, 0.2, 0.4, 0.6\} and we report the best result from them. For CTM, we used the best-performing sentence transformer “all-mpnet-base-v2”\footnote{https://www.sbert.net/docs/pretrained\_models.html} to convert each document into a document embedding. We trained it over 100 epochs for the 20NG, TagMynews and Twitter datasets and 2 epochs for the Dbpedia dataset. We tuned the \textit{n\_sample} parameter and found that setting it to 40 gives the best results. For CWTM, we built our model on top of ``BERT-base-uncased'' and we set the epoch number to 20 for the 20NG, TagMynews and Twitter datasets and set it to 1 for the DBpedia dataset. The Dirichlet prior was to 0.1 and the
e batch size was set to 16. The learning rate was set to 1e-3. We also applied a linear scheduler with 10\% warmup steps for CWTM. For the soft prompt we used, we set the length of the prompt to 10. We trained each model three times to report the average results. All experiences were conducted with NVIDIA GeForce RTX 3090.

\section*{Appendix B: Document Classification Results \labeltext{B}{sec:appendix_B}}
\setcounter{figure}{0}
\renewcommand{\thefigure}{B\arabic{figure}}

\begin{figure}[h!]
\centering
\resizebox{0.48\textwidth}{!}{
\includegraphics[width=1\textwidth]{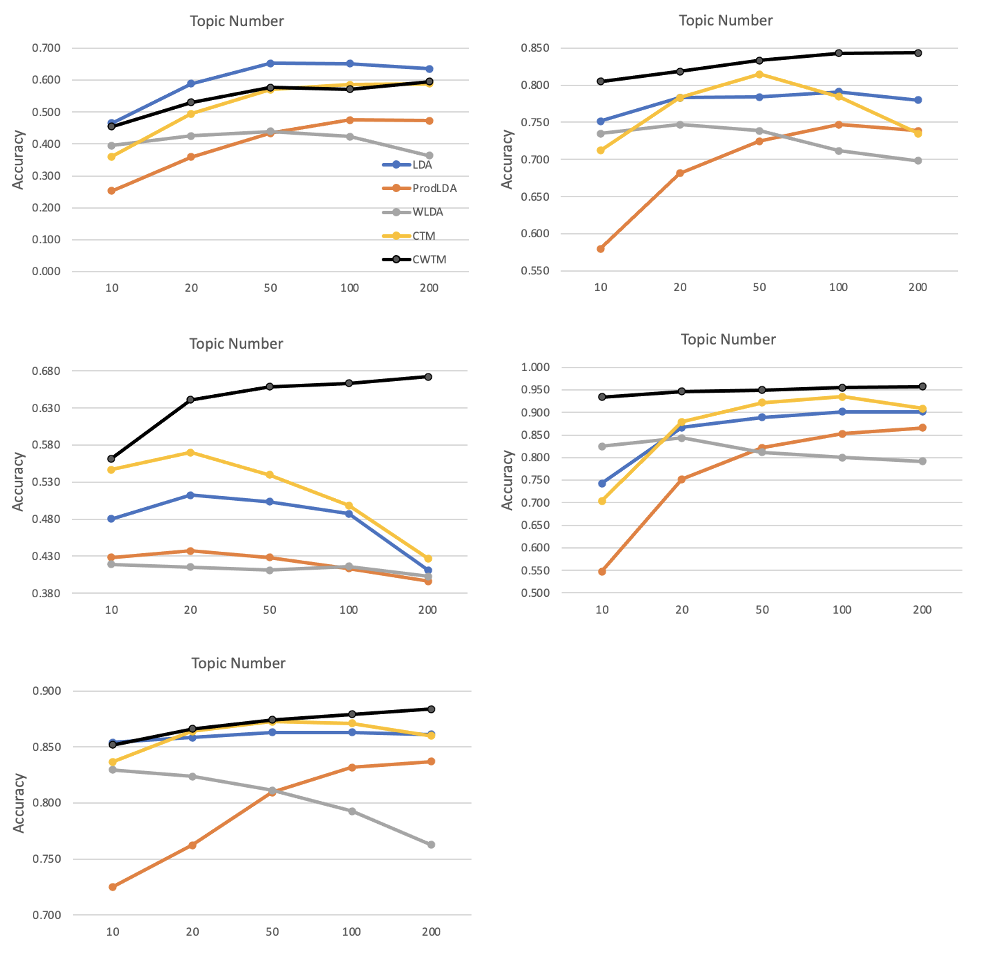}
}
\caption{Document classification accuracy across different numbers of topics by different models for 20NG (top row left); TagMyNews (top row right), Twitter (2nd row left), DBpedia (2nd row right), and AGNews (bottom row).}
\label{Fig.cls}
\end{figure}

We plot the classification performance of different models with different numbers of topics settings in Figure \ref{Fig.cls}. The number of topics is set to $Z$= \{10, 20, 50, 100, 200\}. CWTM achieved better performance than the other models on the TagMyNews, Twitter, Dbpedia, and AGNews datasets. Although LDA outperforms CWTM on the 20NG dataset, CWTM has better performance than the other neural topic model baselines. 

\clearpage
\onecolumn
\section*{Appendix C: Topic Words from CWTM on Different Datasets \labeltext{C}{sec:appendix_C}}

\begin{table*}[h!]
\centering
\resizebox{\textwidth}{!}{%
\begin{tabular}{|l|l|}
\hline
Topic    & Top 10 words                                                                         \\ \hline
topic 1  & key, chip, radio, circuit, phone,   power, battery, encryption, signal, line         \\ \hline
topic 2  & car, bike, engine, mile, brake,   tire, wheel, rear, dealer, bmw                     \\ \hline
topic 3  & god, christian, bible, jesus,   scripture, christ, church, sin, lord, holy           \\ \hline
topic 4  & space, nasa, orbit, doctor,   patient, disease, medical, launch, drug, shuttle       \\ \hline
topic 5  & game, player, hockey, team,   baseball, playoff, pitcher, nhl, play, goal            \\ \hline
topic 6  & drive, card, driver, disk, bus,   memory, ram, controller, mac, machine              \\ \hline
topic 7  & argument, agree, moral, opinion,   discussion, question, point, assume, belief, fact \\ \hline
topic 8  & armenian, gun, child, fbi, weapon,   police, arab, government, law, jew              \\ \hline
topic 9  & window, program, application,   post, unix, code, software, command, server, run     \\ \hline
topic 10 & test, david, ditto, stuff, cheer,   hey, deleted, tony, sex, michael                 \\ \hline
\end{tabular}%
}
\caption{Topic words from CWTM on the 20NG dataset.}
\label{tab:20NG_topics}
\end{table*}

\begin{table*}[h!]
\centering
\resizebox{\textwidth}{!}{%
\begin{tabular}{|l|l|}
\hline
Topic    & Top 10 words                                                                          \\ \hline
topic 1  & nuclear, earthquake, tsunami,   plant, tornado, quake, radiation, power, oil, reactor \\ \hline
topic 2  & stock, bank, billion, investor,   market, fund, price, financial, share, debt         \\ \hline
topic 3  & yankee, inning, sox, phillies,   mets, run, hitter, pitcher, jay, baseball            \\ \hline
topic 4  & broadway, theater, musical, idol,   tony, revival, play, show, stage, star            \\ \hline
topic 5  & study, cancer, risk, drug,   diabetes, patient, disease, woman, heart, treatment      \\ \hline
topic 6  & open, round, federer, murray,   nadal, french, andy, djokovic, final, master          \\ \hline
topic 7  & trial, court, case, police,   accused, charge, prosecutor, judge, charged, guilty     \\ \hline
topic 8  & apple, google, microsoft, android,   tablet, phone, ipad, browser, service, window    \\ \hline
topic 9  & connecticut, east, kentucky, ncaa,   butler, ohio, duke, notre, big, texas            \\ \hline
topic 10 & rebel, libyan, force, libya,   gaddafi, nato, government, japan, military, syrian     \\ \hline
\end{tabular}%
}
\caption{Topic words from CWTM on the TagMyNews dataset.}
\label{tab:TagMyNews_topics}
\end{table*}

\begin{table*}[h!]
\centering
\resizebox{\textwidth}{!}{%
\begin{tabular}{|l|l|}
\hline
Topic    & Top 10 words                                                                                  \\ \hline
topic 1  & customer, twitter, ur, tweet,   email, account, post, awe, service, app                       \\ \hline
topic 2  & fucking, fuck, shit, guy,   hilarious, hell, pout, man, funny, team                           \\ \hline
topic 3  & terrorism, anger, outrage, terror,   black, rage, pakistan, angry, fuming, trump              \\ \hline
topic 4  & amazing, blue, lively, news,   music, rock, musically, tonight, great, free                   \\ \hline
topic 5  & hilarious, horrible, fuming, fury,   depressing, rage, furious, bitter, irritate, nightmare   \\ \hline
topic 6  & good, happy, lost, year, work,   back, night, home, tomorrow, wait                            \\ \hline
topic 7  & depression, fear, sadness,   anxiety, afraid, terrible, bad, worst, hate, cry                 \\ \hline
topic 8  & today, phone, guy, year, work,   service, back, thing, im, call                               \\ \hline
topic 9  & live, broadcast, ly, gbbo, lol,   birthday, gt, snap, wow, snapchat                           \\ \hline
topic 10 & watch, musically, blue, horror,   nightmare, absolutely, delight, exhilarating, glee, playing \\ \hline
\end{tabular}%
}
\caption{Topic words from CWTM on the Twitter dataset.}
\label{tab:Twitter_topics}
\end{table*}

\begin{table*}[]
\centering
\resizebox{\textwidth}{!}{%
\begin{tabular}{|l|l|}
\hline
Topic    & Top 10 words                                                                                     \\ \hline
topic 1  & aircraft, car, engine, wing, seat,   locomotive, fighter, model, motorcycle, air                 \\ \hline
topic 2  & journal, university, newspaper,   published, editor, peer, reviewed, academic, college, magazine \\ \hline
topic 3  & high, public, grade, student,   pennsylvania, township, secondary, education, historic, house    \\ \hline
topic 4  & football, played, footballer,   player, play, professional, league, hockey, playing, team        \\ \hline
topic 5  & directed, starring, drama, book,   movie, comedy, star, story, written, thriller                 \\ \hline
topic 6  & navy, ship, class, launched, hm,   commissioned, submarine, laid, vessel, built                  \\ \hline
topic 7  & band, record, studio, rock, song,   released, singer, track, single, ep                          \\ \hline
topic 8  & mountain, peak, river, range,   lake, mount, alp, elevation, summit, flow                        \\ \hline
topic 9  & plant, found, genus, moth,   endemic, habitat, flowering, native, tree, grows                    \\ \hline
topic 10 & persian, iran, romanized, rural,   gmina, poland, population, province, village, voivodeship     \\ \hline
\end{tabular}%
}
\caption{Topic words from CWTM on the Dbpedia dataset.}
\label{tab:Dbpedia_topics}
\end{table*}

\begin{table*}[h!]
\centering 
\resizebox{\textwidth}{!}{%
\begin{tabular}{|l|l|}
\hline
Topic    & Top 10 words                                                                                   \\ \hline
topic 1  & stock, price, dollar, market,   investor, higher, future, high, rise, rose                     \\ \hline
topic 2  & court, charge, case, police,   arrested, charged, lawsuit, judge, trial, drug                  \\ \hline
topic 3  & oil, space, moon, spacecraft,   crude, nasa, titan, station, saturn, planet                    \\ \hline
topic 4  & arsenal, goal, liverpool, chelsea,   united, manchester, striker, england, test, champion      \\ \hline
topic 5  & corp, million, billion, deal, buy,   bid, group, business, takeover, share                     \\ \hline
topic 6  & inning, sox, yankee, league,   giant, astros, dodger, cub, twin, nl                            \\ \hline
topic 7  & music, game, store, video, dvd,   nintendo, sony, make, song, ipod                             \\ \hline
topic 8  & baghdad, militant, iraqi, iraq,   insurgent, rebel, killed, soldier, attack, troop             \\ \hline
topic 9  & microsoft, linux, software,   version, window, source, server, system, application, enterprise \\ \hline
topic 10 & open, round, final, championship,   master, cup, champion, federer, seed, hewitt               \\ \hline
\end{tabular}%
}
\caption{Topic words from CWTM on the AGNews dataset.}
\label{tab:Dbpedia_topics}
\end{table*}

\end{document}